\documentclass[10pt,twocolumn,letterpaper]{article}

\usepackage[pagenumbers]{cvpr} 

\definecolor{cvprblue}{rgb}{0.21,0.49,0.74}
\usepackage[pagebackref,breaklinks,colorlinks,allcolors=cvprblue]{hyperref}

\usepackage{multirow}
\usepackage{makecell}
\usepackage{pifont}
\usepackage{arydshln}
\usepackage{booktabs}
\usepackage{balance}
\usepackage{subcaption}
\usepackage{graphicx}
\usepackage{marvosym}
\usepackage{threeparttable}
\usepackage{multirow}
\usepackage{colortbl}
\usepackage{array}
\usepackage{ragged2e}
\usepackage{tabularx}
\usepackage[table]{xcolor} 

\def\paperID{12083} 
\def\confName{CVPR}
\def\confYear{2026}

\title{SkeletonContext: Skeleton-side Context Prompt Learning for Zero-Shot \\ Skeleton-based Action Recognition}

\vspace{-20pt}

\author{Ning Wang$^{1,2}$\thanks{Part of this work was completed during the Ning's visit to The University of Western Australia.}, Tieyue Wu$^{2}$, Naeha Sharif$^{3}$, Farid Boussaid$^{3}$, Guangming Zhu$^{2}$,  \\ Lin Mei$^{4}$, Mohammed Bennamoun$^{3}$, Liang Zhang$^{2}$\thanks{Corresponding author.}\\
    $^{1}$Chang'an University,$^{2}$Xidian University, $^{3}$University of Western Australia,$^{4}$Donghai Lab\\
    {\tt\small ning@chd.edu.cn, \quad gxyzd648@gmail.com, \quad \{gmzhu, liangzhang\}@xidian.edu.cn,} \\
    {\tt\small \{naeha.sharif,farid.boussaid,mohammed.bennamoun\}@uwa.edu.au, \quad meilin@donghailab.com}
}

\begin{document}
\maketitle
\begin{abstract}
Zero-shot skeleton-based action recognition aims to recognize unseen actions by transferring knowledge from seen categories through semantic descriptions. Most existing methods typically align skeleton features with textual embeddings within a shared latent space. However, the absence of contextual cues, such as objects involved in the action, introduces an inherent gap between skeleton and semantic representations, making it difficult to distinguish visually similar actions.
To address this, we propose SkeletonContext, a prompt-based framework that enriches skeletal motion representations with language-driven contextual semantics.
Specifically, we introduce a Cross-Modal Context Prompt Module, which leverages a pretrained language model to reconstruct masked contextual prompts under guidance derived from LLMs.
This design effectively transfers linguistic context to the skeleton encoder for instance-level semantic grounding and improved cross-modal alignment.
In addition, a Key-Part Decoupling Module is incorporated to decouple motion-relevant joint features, ensuring robust action understanding even in the absence of explicit object interactions.
Extensive experiments on multiple benchmarks demonstrate that SkeletonContext achieves state-of-the-art performance under both conventional and generalized zero-shot settings, validating its effectiveness in reasoning about context and distinguishing fine-grained, visually similar actions. Our project is available at \url{https://github.com/NingWang2049/skeletoncontext}.
\end{abstract}
\vspace{-20pt}    
\section{Introduction}
\label{sec:intro}
\begin{figure}
    \centering
    \includegraphics[width=0.9\linewidth]{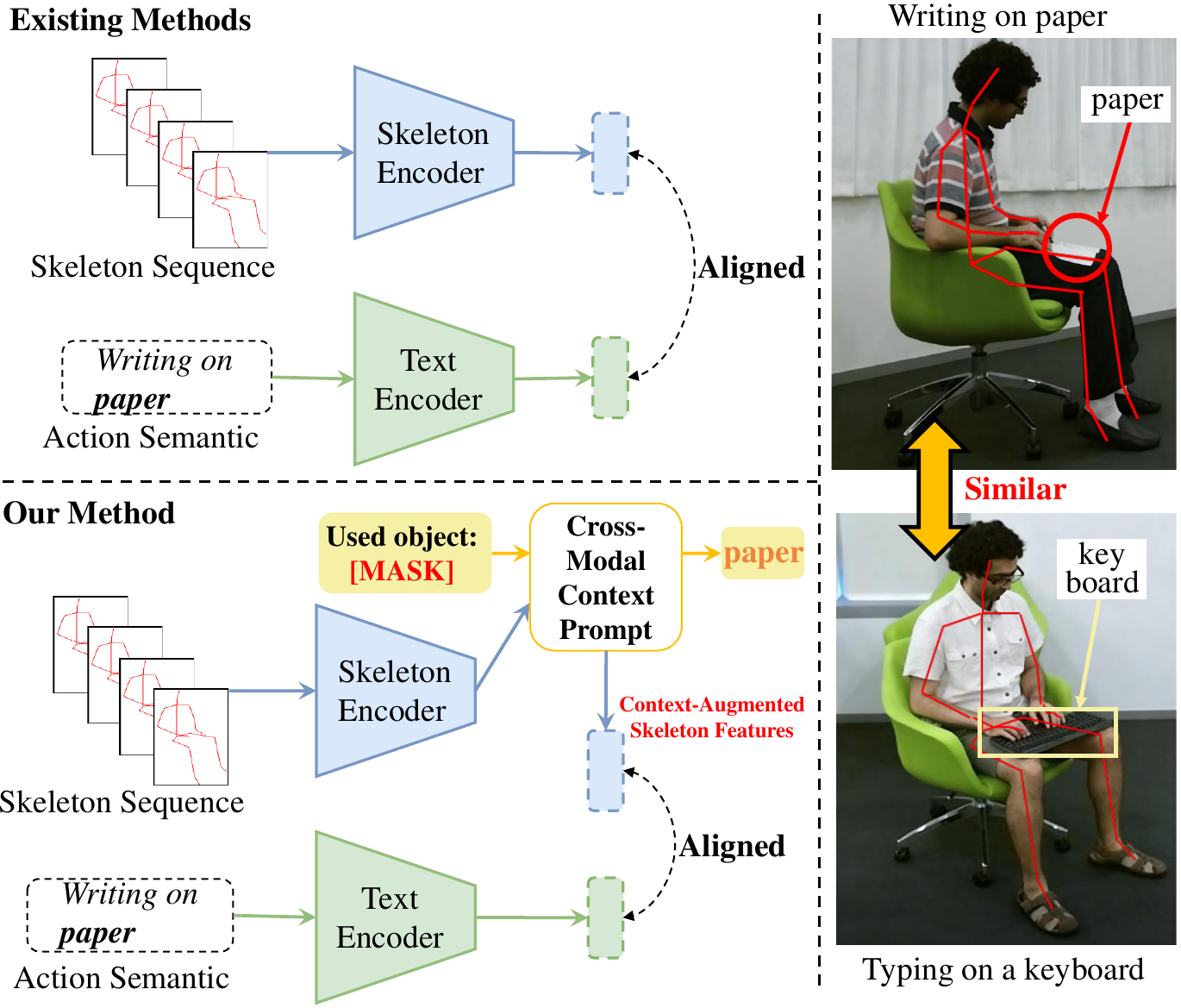}
    \caption{\textbf{Comparison between existing ZSSAR methods and our proposed SkeletonContext.}
Conventional ZSSAR methods directly align skeleton features with textual descriptions, but the absence of contextual cues creates a semantic gap that hinders discrimination between similar actions. In contrast, SkeletonContext reconstructs language-driven contextual semantics (e.g., objects) and injects them into skeleton representations, enabling fine-grained, context-aware zero-shot action recognition.}
    \label{fig:intro}
    \vspace{-0.3in}
\end{figure}


Human action recognition plays a crucial role in numerous applications, including human-computer interaction~\cite{nayak2021human, kashef2021smart}, intelligent surveillance~\cite{singh2019multi, shorfuzzaman2021towards}, and healthcare monitoring~\cite{waheed2020automatic, zhang2020human}. 
Skeleton-based representations~\cite{zhu2023motionbert,yang2024one,yan2024crossglg,liu2025adaptive}, which model human motion through joint coordinates, are valuable in privacy-sensitive and real-world scenarios due to their robustness to appearance variations, background clutter, and view changes. 
Most existing skeleton-based action recognition methods~\cite{chi2022infogcn, chen2021channel, liu2023transkeleton, xin2023skeleton, wu2025frequency} rely on fully supervised learning with pre-defined categories, limiting their performance to diverse and unseen action categories. 
To overcome this limitation, zero-shot skeleton-based action recognition (ZSSAR) has emerged, enabling models to recognize unseen actions using knowledge from seen action categories and semantic descriptions.

Existing ZSSAR approaches~\cite{hubert2017learning, gupta2021syntactically, zhou2023zero,li2023multi,chen2024fine, zhu2024part,li2024sa,zhu2025semantic} align skeleton and text embeddings within a shared latent space to enhance cross-modal generalization.
They mainly focus on leveraging additional external knowledge \cite{chen2024fine,zhu2024part}, augmenting training data \cite{gupta2021syntactically, zhu2025semantic}, or designing more powerful skeleton encoders \cite{zhou2023zero,li2023multi, li2024sa}.
However, skeleton sequences, unlike raw videos, lack detailed appearance and contextual cues, such as the keyboard in the action category “typing on a keyboard” shown in Fig. \ref{fig:intro}.
As a result, the lack of contextual cues in skeleton data inherently hampers its alignment with semantic representations, limiting the model’s ability to capture fine-grained semantic distinctions between similar actions, such as “typing on a keyboard” and “writing on paper”.
Although fully supervised fine-grained skeleton-based action recognition methods\cite{zhou2023learning,liu2025revealing} have explored learning discriminative representations to distinguish ambiguous actions, they rely heavily on class-specific supervision and contrastive calibration between known categories. 
As a result, these methods cannot be directly extended to zero-shot settings, where unseen classes lack labeled instances and discriminative prototypes \cite{liu2025revealing} cannot be established.

To address this, we argue that enhancing skeleton representations by reconstructing contextual semantics, such as associated objects, is crucial for improving cross-modal alignment and distinguishing visually similar actions.
In this paper, we propose SkeletonContext, a prompt-based framework that enriches skeletal motion representations with language-driven contextual knowledge.
As illustrated in Fig.~\ref{fig:intro}, our framework injects reconstructed contextual semantics into the skeleton modality, achieving more reliable alignment with textual semantics.
Specifically, we first extract action-related contextual descriptions from large language models (LLMs), capturing external knowledge such as interacted-with objects and environments that are absent in the skeleton modality.
Then, our Cross-Modal Context Prompt Module reconstructs masked contextual prompts under the supervision of generated semantics (e.g., objects and environments). This reconstruction process transfers contextual knowledge from a pretrained language model, resulting in context-enhanced skeleton representations.
We further enhance the Cross-Modal Context Prompt Module by designing a differential joint encoder to model fine-grained motion dynamics across joints, and a progressive partial masking strategy to progressively refine contextual learning by encouraging robustness under incomplete motion cues.
Different from prior prompt-based ZSSAR methods (e.g., SCoPLe \cite{zhu2025semantic}, Neuron \cite{chen2024neuron}) that enhance text encoders for better semantic alignment, SkeletonContext enriches the skeleton encoder itself with language-driven contextual prompts, thus fundamentally improving motion-side representation and cross-modal grounding.
Moreover, to handle actions without explicit object interactions, we also propose a Key Part Decoupling Module that disentangles motion-related joint features and highlights the most relevant body parts.

\noindent \textbf{To summarize, our contributions are three-fold}: 
\begin{itemize}
    \item We propose SkeletonContext, a novel context-aware framework for zero-shot skeleton-based action recognition. It introduces language-driven contextual knowledge into the skeleton modality, effectively bridging the semantic gap between motion and language.
    \item We demonstrate that injecting contextual semantics into skeleton representations via a cross-modal prompt learning mechanism significantly enhances the model’s ability to distinguish visually similar actions and generalize to unseen categories.
    \item We conduct extensive experiments on three benchmark datasets. SkeletonContext consistently achieves state-of-the-art performance under both conventional and generalized zero-shot settings, validating the effectiveness and robustness of our approach.
\end{itemize}
\vspace{-0.1in}
\section{Related Work}
\label{sec:relatedwork}
\subsection{Zero-Shot Skeleton Action Recognition}
Zero-shot skeleton action recognition (ZSSAR) seeks to recognize unseen action categories by transferring knowledge from seen classes through shared semantic embeddings. Early studies focused on global alignment between skeleton representations and textual semantics. RelationNet~\cite{jasani2019skeleton} introduced metric-based embedding alignment, while SynSE \cite{gupta2021syntactically} employed a generative VAE to couple global skeleton features with verb–noun category embeddings. SMIE \cite{zhou2023zero} further imposed temporal mutual-information constraints to enhance global correspondence. However, these approaches relied on hand-crafted category names and coarse global mappings, which limited semantic granularity and discriminative power.
To improve fine-grained reasoning, PURLS \cite{zhu2024part} and STAR \cite{chen2024fine} decomposed skeletons into overlapping parts, using LLM-generated part descriptions for local alignment. 
SA-DAVE \cite{li2024sa} addressed this by decoupling semantic-relevant and irrelevant skeleton components, improving generalization but still performing static alignment on global semantics.

Recent research has advanced ZSSAR by enhancing semantic diversity, contextual awareness, and dynamic alignment. Neuron \cite{chen2024neuron} proposed a dynamically evolving dual skeleton-semantic synergistic framework guided by multi-turn LLM-generated side information, enabling controlled micro-to-macro cross-modal alignment via spatial compression and temporal memory mechanisms. FS-VAE \cite{wu2025frequency} introduced frequency-semantic modeling, decomposing skeleton motions into low- and high-frequency components to enrich semantic features and applying a calibrated cross-alignment loss to mitigate noisy skeleton–text pairs. Meanwhile, SCoPLe \cite{zhu2025semantic} proposed semantic-guided cross-modal prompt learning that jointly tunes textual and skeletal prompts for fully data-driven semantic alignment, eliminating handcrafted descriptions and achieving state-of-the-art zero-shot performance.

\subsection{Contextual Prompt Learning}
Before the advent of prompting, fine-tuning was the dominant strategy for adapting pre-trained large models (PLMs) to downstream tasks \cite{liu2023pre}. 
Although effective, it required updating a massive number of parameters and incurred substantial computational and storage costs, especially as model sizes grew from millions to trillions of parameters \cite{fedus2022switch,lewis2020bart,radford2019language}. 
To overcome this inefficiency, researchers introduced lightweight adaptation methods that freeze most pre-trained weights while learning a small set of additional parameters. 
Among them, prompt learning emerged as a particularly efficient and flexible approach.
Early works~\cite{su2019vl,tan2019lxmert,chen2020uniter} in the vision–language community employed masked reconstruction objectives to align textual and visual representations. 
These models learned to predict masked words or image regions from cross-modal context, effectively enabling implicit prompt-like conditioning during training. 
Such masked multimodal pre-training frameworks demonstrated that recovering missing semantics from one modality using cues from another could transfer rich contextual knowledge, revealing that prompts can act as bridges for cross-modal reasoning.
Building on this foundation, recent visual prompt tuning \cite{jia2022visual} and conditional prompt generation method \cite{zhou2022conditional} extended these ideas to vision and multimodal models (e.g., CLIP-based frameworks \cite{khattak2023maple}), improving generalization across unseen categories. 

However, these methods assume that contextual information, e.g., objects, textures, or environments, is visually present and directly observable. 
This assumption breaks down in skeleton-based action recognition, where the input consists solely of abstract joint coordinates.
To address this limitation, we introduce a reconstruction-based prompting strategy that leverages a pretrained language model to recover masked contextual slots from skeletal motion. 
This enables language-driven cues to enrich skeleton representations, transferring contextual semantics that are inherently missing in the visual modality.

\begin{figure*}[htb]
    \centering
    \includegraphics[width=0.8\linewidth]{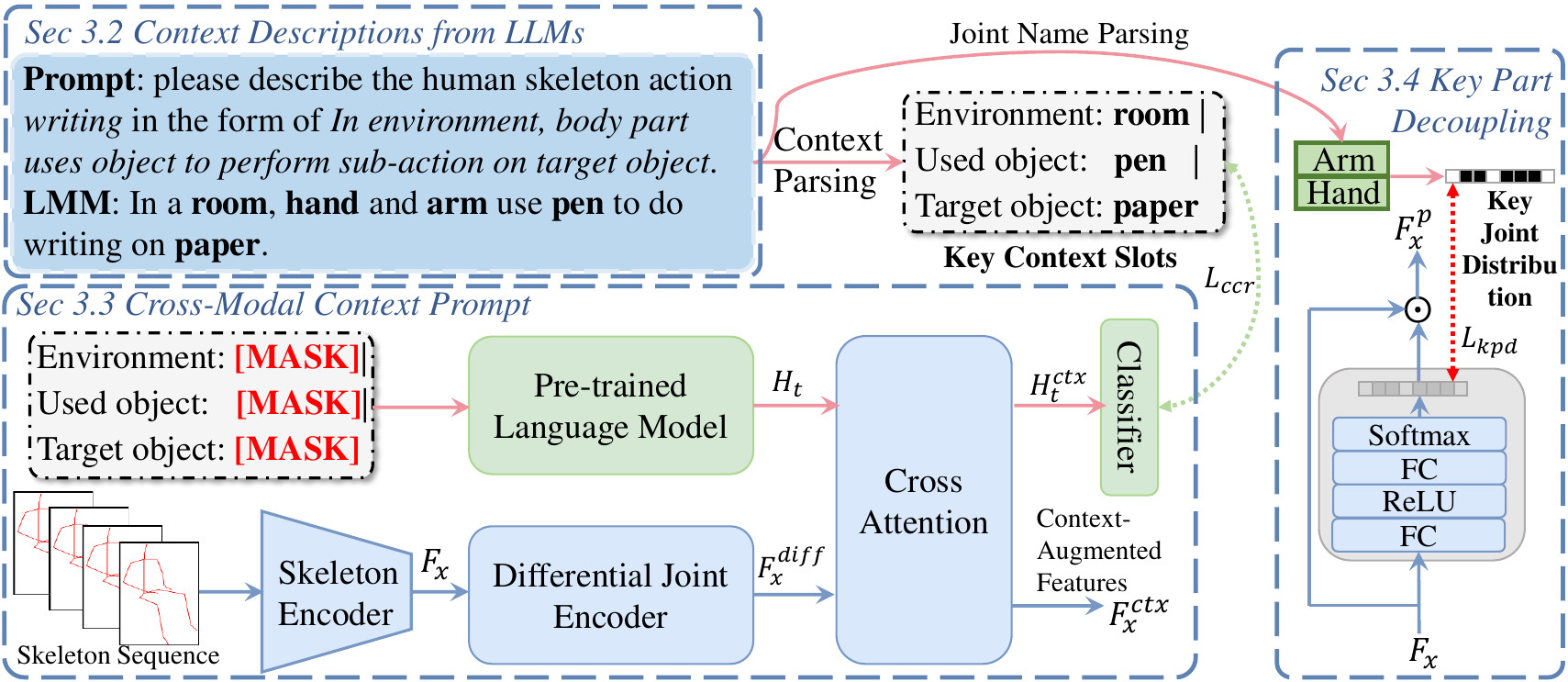}
    \caption{\textbf{Overview of SkeletonContext.} The model enriches skeleton features via Cross-Modal Context Prompt Module (Sec \ref{sec:context_reconstruction_module}) and Key-Part Decoupling Module (Sec \ref{sec:keypart}) for context-aware zero-shot action recognition.
    Both modules are guided by LLM–derived contextual knowledge  (Sec \ref{sec:context_descriptions}) during training, enabling semantic grounding and cross-modal alignment.}
    \label{fig:method}
    \vspace{-0.2in}
\end{figure*}

\section{Method}
\subsection{Problem Definition}
Let $\mathcal{D}$ be a skeleton-based action dataset composed of a set of skeleton sequences $\mathcal{X}$ and a corresponding label set $\mathcal{Y}$, in which $y \in \mathcal{Y}$ is the corresponding action label. 
The dataset is partitioned into two disjoint subsets: the seen set $\mathcal{X}_s$ and the unseen set $\mathcal{X}_u$. 
The goal of zero-shot skeleton-based action recognition is to learn a model using only $\mathcal{X}_s$ and $\mathcal{Y}_s$ that can correctly classify samples $x \in \mathcal{X}_u$ into their corresponding unseen label in $\mathcal{Y}_u$.
There are two evaluation protocols commonly used in this task:
(a) Zero-Shot Learning (ZSL): The model is evaluated only on the unseen classes, i.e., predicting $x \in \mathcal{X}_u$ among $\mathcal{Y}_u$.
(b) Generalized Zero-Shot Learning (GZSL): The model is evaluated on both seen and unseen classes, i.e., predicting $x \in \mathcal{X}_u \cup \mathcal{X}_s$ among $\mathcal{Y}_u \cup \mathcal{Y}_s$.
A common baseline approach is to train a skeleton encoder on the seen classes using standard cross-entropy loss. This model extracts spatial-temporal features from skeleton sequences, denoted as $F_x = \Phi(x)$, where $\Phi(\cdot)$ is the skeleton encoder (e.g., Shift-GCN). For semantic representation, a semantic encoder is employed to encode class label or textual description $y$ into semantic embeddings, denoted as $Z_y = \Psi(y)$, where $\Psi(\cdot)$ is the semantic encoder (e.g., CLIP).
The challenge lies in aligning $F_x$ and $Z_y$ in a shared latent space that enables effective cross-modal knowledge transfer between seen and unseen action categories. 
Our method addresses this problem by reconstructing the missing context information of $F_x$, and aligning the reconstructed $F_x^c$ with $Z_y$ to enhance the generalization ability.

\subsection{Context Descriptions from LLMs}
\label{sec:context_descriptions}
Previous studies \cite{xiang2023generative,zhu2024part,li2024sa} typically utilize action category names or enriched action descriptions as the semantic representations. 
However, such formulations tend to overlook the external context in which actions are performed.
In this paper, we generate the structured semantic descriptions that incorporate contextual information beyond the skeleton modality.
Specifically, we leverage LLMs to generate compositional descriptions that capture key elements such as interaction objects, environmental settings, and body part involvement.
To elicit such descriptions, we design a prompt template that guides the LLM to generate context-aware descriptions.
For example, we obtain the description \textit{In office, hand uses pen to write on paper} of action "writing" with the question: \textit{please describe the human skeleton action \texttt{writing} in the form of In \texttt{environment}, \texttt{body part} uses \texttt{object} to perform \texttt{subaction} on \texttt{target object}}.
Meanwhile, we allow contextual elements to appear in multiple action descriptions. For example, the object “cup” may appear in both “drinking water” and “pouring coffee”, thereby facilitating the transfer of contextual knowledge between different categories.
Furthermore, to capture intra-class diversity, we generate multiple context descriptions for each action category. 
This enriches the semantic space with diverse yet transferability representations, improving the robustness of cross-modal alignment.

\subsection{Cross-Modal Context Prompt Module}
\label{sec:context_reconstruction_module}
To inject the missing context semantics into original skeleton sequences, we propose a cross-modal context prompt module.
This module aims to (i) reconstruct contextual textual descriptions of an action from a masked prompt, and (ii) utilize the reconstructed language as cross-modal supervision to enhance skeleton representations for ZSSAR.
It is instantiated on top of a pre-trained language model and a language-skeleton interaction layer, enabling bidirectional information flow between token-level linguistic representations and skeleton features.
This design follows the insight that cross-modal masked modeling, which encourages models to ground ambiguous tokens in the complementary modality, has proven effective in vision–language pretraining frameworks \cite{bao2022vl,chen2023duet,li2020oscar}.

\subsubsection{Differential Joint Encoder}
To capture subtle motion differences and model fine-grained joint dependencies, we propose a Differential Joint Encoder (DJE) to encode skeletal features.
The key idea is to capture the inter-joint differences to highlight pose-specific dependencies that implicitly reflect contextual clues (e.g., "leaning over a table" is associated with a scene similar to writing).
In practice, we start by pooling the skeleton feature $F_x$ to the topology-level skeleton feature $F_x' \in \mathbb{R}^{N \times C}$.
Subsequently, we normalize and map $F_x'$ to query and key  with learnable matrices $W^Q, W^K$.
The projected features could be formulated as $H_x^Q = \text{LN}(F_x')W_Q^\top$ and $H_x^K = \text{LN}(F_x')W_K^\top$, where $\text{LN}$ denotes the Layer Normalization \cite{ba2016layer}.
Then, for samples with different postures, the differences between samples can be captured through pairwise comparisons between joints.
Specifically, we compute a differential topology representation $A^{diff} = \phi(\mathcal{T}_1(H_x^Q) - \mathcal{T}_2(H_x^K))$, where $\phi$ represents the activation function, and $\mathcal{T}_d$ expand the tensor at its $d$-th dimension and replicate the result $N$ times along that dimension.
The topology-enhanced embedding is then updated as 
\begin{equation}
    F_x^{diff} = \text{Avg}(A^{diff} \cdot \text{LN}(F_x')W_V^\top) + F_x'
\end{equation}
This operation highlights discriminative spatial dependencies that often correlate with specific environmental or interaction contexts, establishing the foundation for context reconstruction.

\subsubsection{Progressive Partial Masking}
Each context-aware descriptions generated in Sec \ref{sec:context_descriptions} includes three contextual slots: \texttt{Environment}, \texttt{Used object}, and \texttt{Target}. 
To guide the model to infer the missing context semantics, we mask these slots in the textual input as a prompt template:
\[
{\footnotesize
\underbrace{\text{Environment: }}_{\text{Prompt}}
\underbrace{\text{[MASK]}}_{\text{Context}}
\mid
\underbrace{\text{Used object: }}_{\text{Prompt}}
\underbrace{\text{[MASK]}}_{\text{Context}} 
\mid
\underbrace{\text{Target object: }}_{\text{Prompt}}
\underbrace{\text{[MASK]}}_{\text{Context}}
}
\]
During training, we randomly sample one candidate description for each action class at every iteration and extract its contextual slots to fill the masked positions in the prompt template, serving as the ground-truth context $C_y$.
However, directly reconstructing all contextual elements (environment, used object, target object) from limited skeleton cues is extremely challenging. 
Moreover, our concise structured prompt differs in form and complexity from the long, natural sentences that pre-trained language models are originally trained on. 
To bridge this distributional gap and stabilize the optimization process, we introduce a \textit{Progressive Partial Masking (PPM)} mechanism that gradually increases the level of masking difficulty over training.

Formally, we define a masking function $\tilde{C}_y = \text{PPM}(r_t, C_y)$, where $r_t = min(1, t/T) \in [0,1]$ denotes the masking ratio at training step $t$. 
At the beginning of training, $r_t$ is set to a small value so that only a small subset of contextual slots is randomly masked, making the reconstruction task easier. 
As training progresses, $r_t$ gradually increases until all contextual elements in $C_y$ are masked, forcing the model to infer the complete context solely from the skeleton representation and linguistic priors. 
This gradual increase in masking ratio serves as a curriculum that transitions the model from partial reconstruction to full-context reasoning, effectively narrowing the domain gap between structured prompts and natural language.

\subsubsection{Semantic Context Grounding}
Let $H_t \in \mathbb{R}^{L \times d}$ denote the hidden states of the contextual prompt fed into BERT \cite{alaparthi2020bidirectional}.
Subsequently, A cross attention layer is used for cross-modal information transfer.
The cross-attention operation is defined as:
\begin{align}
    F_x^{ctx} = \text{CrossAttn}_{x \rightarrow t}(F_x^{diff}, H_t), \\
    H_t^{ctx} = \text{CrossAttn}_{t \rightarrow x}(H_t, F_x^{diff}), 
\end{align}
where $\text{CrossAttn}(\cdot)$ consists of one bi-directional cross-attention block, two self-attention blocks \cite{vaswani2017attention} and two feed-forward blocks.
A residual connection and layer normalization \cite{ba2016layer} are added to each block.
The keys and values from each modality are passed as the input to other modality’s multi-headed attention blocks.
Let $\hat{C_x} = f_{\text{MLM}}(H_t^{ctx})$ be the masked target context items, where text decoder $f_{\text{MLM}}(\cdot)$ transforms the contextualized token representations into word prediction probabilities. The objective is
\begin{equation}
     \mathcal{L}_{ccr} = - \sum_{i=1}^{Len(\tilde{C}_y)} log P(\hat{C}_x^i|\tilde{C}_y^i) ,
\end{equation}
where 
\begin{equation}
   P(\hat{C}_x^i|\tilde{C}_y^i) = - log(\frac{exp(\hat{C}_x^i \cdot \tilde{C}_y^i)}{\sum_{C' \in \mathcal{V}}exp(\hat{C}_x^i \cdot \tilde{C}_y^i)}).
\end{equation}
This loss encourages the model to reconstruct plausible contextual semantics guided jointly by skeleton motion patterns and the linguistic priors encoded in the language model.
During inference, the model infers missing contextual information from motion cues and the linguistic knowledge encoded in BERT \cite{alaparthi2020bidirectional}, enabling reasoning across skeleton and language modalities.

\subsection{Key Part Decoupling Module}
\label{sec:keypart}
Certain actions (e.g., \textit{hand waving}) are primarily defined by local motion patters rather than external context. 
To better model such context-independent actions, we further propose the Key Part Decoupling (KPD) module that emphasizes discriminative joints.
Specifically, given $F_x$, it predicts a joint-importance map $K_{\text{out}}$ through two linear layers and a softmax function, generating reweighted features:
\begin{equation}
F_x^p = K_{\text{out}} \odot F_x,
\end{equation}
where $\odot$ denotes element-wise multiplication. 
We derive a prior importance distribution $K_{\text{gt}}  \in [0,1]^{T \times V}$ from the \texttt{[body part]} field and impose a calibration loss:
\begin{equation}
\mathcal{L}_{\text{kpd}} = \sum_{t=1}^{T} \| K_{\text{out},t} - K_{\text{gt}} \|_2.
\end{equation}
Although the guidance for KPD comes from seen classes, the disentanglement it learns is context-driven rather than class-dependent. The module captures general motion–context relations (e.g., hand–object or head–body interactions), which naturally transfer to unseen actions that share similar structural patterns.

\subsection{Training and Inference}
\subsubsection{Training}
To align the skeleton and semantic spaces, we compute the dot product between the pooled context-augmented skeleton features $\hat{F}_x^{ctx}$ and the pooled semantic embeddings $\hat{Z}_y^c$, and optimize a contrastive cross-entropy loss to encourage matched skeleton–semantic pairs to have higher similarity.
Similarly, the pooled part-level features $\hat{F}_x^{p}$ are aligned with the action-aware semantic embeddings $\hat{Z}_y^a$ ~\cite{chen2024neuron} to capture motion-relevant semantics.
We define the alignment operation as:
\begin{align}
    \mathcal{L}_{align} = -\log\frac{\exp\!\big(\phi_c(\hat{F}_x^{ctx})^\top \psi(\hat{Z}_y^{c})\big)}{\sum_{k\in Y_s}\exp\!\big(\phi_c(\hat{F}_x^{ctx})^\top \psi(\hat{Z}_k^{c})\big)} \\
    - \log\frac{\exp\!\big(\phi_p(\hat{F}_x^{p})^\top \psi(\hat{Z}_y^{p})\big)}{\sum_{k\in Y_s}\exp\!\big(\phi_p(\hat{F}_x^{p})^\top \psi(\hat{Z}_k^{p})\big)}, 
\end{align}
where $\phi_c(\cdot)$, $\phi_p(\cdot)$, and $\phi(\cdot)$ are project functions. Last, the overall optimization objective is defined as:
\begin{equation}
\mathcal{L}=\mathcal{L}_{align}+\mathcal{L}_{\text{ccr}}+\mathcal{L}_{\text{kpd}}.
\end{equation}

\subsubsection{Inference}
At inference, we predict unseen classes by computing similarities to unseen semantics. To mitigate domain shift in GZSL, we adopt calibrated stacking:
\begin{equation}
\hat{y}_c = \arg\max_{y\in Y_u / Y}\ \rho_s(\tilde{F}_x^{ctx})^\top \phi(\hat{Z}_y^{c}) - \gamma_s\,\mathbb{I}[y\in Y_s], 
\end{equation}
\begin{equation}
\hat{y}_p = \arg\max_{y\in Y_u / Y}\ \rho_t(\tilde{F}_x^{p})^\top \phi(\hat{Z}_y^{p}) - \gamma_t\,\mathbb{I}[y\in Y_s], 
\end{equation}
where $Y_u / Y$ denotes the ZSL/GZSL setting.
Following prior work~\cite{chen2024neuron}, we adopt the same calibration factors $\gamma_s$ and $\gamma_t$ setting when performing GZSL prediction.
The final prediction is then formed by aggregating the context- and part-level estimates as $\hat{y}^\ast=\{\hat{y}_c,\hat{y}_p\}$.
\vspace{-0.1in}
\begin{table*}[htbp]
\scriptsize
\renewcommand\arraystretch{1.0}
\centering
\vspace{-10pt}
\caption{\textbf{Comparison of generalized zero-shot action recognition accuracy on the NTU RGB+D datasets.} SkeletonContext achieves competitive or superior harmonic-mean (H) performance across most evaluation splits, indicating balanced generalization from seen (S) to unseen (U) classes.}
\vspace{-5pt}
\resizebox{1.0\textwidth}{!}{
\begin{tabular}{c|c|c|c|c|c|c|c|c|c|c|c|c|c}
\hline
\multirow{2}{*}{Methods} & \multirow{2}{*}{Venue} & \multicolumn{3}{c|}{NTU-60 (55/5 split)} & \multicolumn{3}{c|}{NTU-60 (48/12 split)} & \multicolumn{3}{c|}{NTU-120 (110/10 split)} & \multicolumn{3}{c}{NTU-120 (96/24 split)} \\ \cline{3-14} 
 &  & Seen & Unseen & H & Seen & Unseen & H & Seen & Unseen & H & Seen & Unseen & H \\ \hline
ReViSE\cite{hubert2017learning} & ICCV 2017 & 74.2 & 34.7 & 29.2 & 62.4 & 20.8 & 31.2 & 48.7 & 44.8 & 46.7 & 49.7 & 25.1 & 33.3 \\
JPoSE\cite{wray2019fine} & ICCV 2019 & 64.4 & 50.3 & 56.5 & 60.5 & 20.6 & 30.8 & 47.7 & 46.4 & 47.0 & 38.6 & 22.8 & 28.7 \\
CADA-VAE\cite{schonfeld2019generalized} & CVPR 2019 & 69.4 & 61.8 & 65.4 & 51.3 & 27.0 & 35.4 & 47.2 & 19.8 & 48.4 & 41.1 & 34.1 & 37.3 \\
SynSE\cite{gupta2021syntactically} & ICIP2021 & 61.3 & 56.9 & 59.0 & 52.2 & 27.9 & 36.3 & 52.5 & 57.6 & 54.9 & 56.4 & 32.2 & 41.0 \\
STAR\cite{chen2024fine} & ACMM2024 & 69.0 & 69.9 & 69.4 & \textcolor{blue}{62.7} & 37.0 & 46.6 & 59.9 & 52.7 & 56.1 & 51.2 & 36.9 & 42.9 \\
SA-DVAE\cite{li2024sa} & ECCV2024 & 62.3 & 70.8 & 66.3 & 50.2 & 36.9 & 42.6 & 61.1 & 59.8 & 60.4 & 58.8 & 35.8 & 44.5 \\ 
ScoPLe\cite{zhu2025semantic} & CVPR2025 & 69.6 & 71.9 & 70.8 & 54.5 & \textcolor{red}{61.8} & 57.9 & \textcolor{blue}{61.1} & 62.3 & 52.2 & 53.3 & 51.2 & 52.2 \\ 
Neuron\cite{chen2024neuron} & CVPR2025 & 69.1 & 73.8 & 71.4 & 61.6 & 56.8 & \textcolor{blue}{59.1} & \textcolor{red}{67.6} & 59.5 & \textcolor{red}{63.3} & \textcolor{red}{67.5} & 44.4 & 53.6 \\ 
FS-VAE\cite{wu2025frequency} & ICCV2025 & \textcolor{blue}{77.0} & \textcolor{blue}{74.5} & \textcolor{blue}{75.7} & 56.2 & 48.6 & 52.1 & 59.2 & \textcolor{red}{67.9} & \textcolor{red}{63.3} & 57.8 & \textcolor{red}{51.9} & \textcolor{blue}{54.7} \\ 
\hline
\textbf{Ours} & \textbackslash{} & 
\textbf{\textcolor{red}{78.3}} & \textbf{\textcolor{red}{76.1}} & \textbf{\textcolor{red}{77.1}} & 
\textbf{\textcolor{red}{65.5}} & \textbf{\textcolor{blue}{57.1}} & \textbf{\textcolor{red}{61.1}} & 
\textbf{\textcolor{blue}{61.1}} &
\textbf{\textcolor{blue}{65.5}} & \textbf{\textcolor{blue}{63.1}} & 
\textbf{\textcolor{blue}{64.6}} & \textbf{\textcolor{blue}{49.6}} & \textbf{\textcolor{red}{56.1}} \\ \hline
\end{tabular}
}
\label{tab:gzsl}
\end{table*}

\begin{table}[t]
\scriptsize
\renewcommand\arraystretch{1.1}
\centering
\vspace{-10pt}
\caption{\textbf{Comparison of zero-shot learning (ZSL) accuracy on the NTU RGB+D datasets.} SkeletonContext achieves strong or comparable results across multiple splits, demonstrating effective generalization to unseen action categories.}
\vspace{-5pt}
\setlength\tabcolsep{3.0pt} 
{
\begin{tabular}{c|c|cc|cc}
\hline
\multirow{2}{*}{Methods} & \multirow{2}{*}{Venue} & \multicolumn{2}{c|}{NTU-60 (ACC,\%)} & \multicolumn{2}{c}{NTU-120 (ACC,\%)} \\ \cline{3-6} 
 &  & \multicolumn{1}{c|}{55/5 split} & 48/12 split & \multicolumn{1}{c|}{110/10 split} & 96/24 split \\ \hline
ReViSE\cite{hubert2017learning} & ICCV2017 & \multicolumn{1}{c|}{53.9} & 17.5 & \multicolumn{1}{c|}{55.0} & 32.4 \\
JPoSE\cite{wray2019fine} & ICCV2019 & \multicolumn{1}{c|}{64.8} & 28.8 & \multicolumn{1}{c|}{51.9} & 32.4 \\
CADA-VAE\cite{schonfeld2019generalized} & CVPR2019 & \multicolumn{1}{c|}{76.8} & 29.0 & \multicolumn{1}{c|}{59.5} & 35.8 \\
SynSE\cite{gupta2021syntactically} & ICIP2021 & \multicolumn{1}{c|}{75.8} & 33.3 & \multicolumn{1}{c|}{62.7} & 38.7 \\
SMIE\cite{zhou2023zero} & ACMM2023 & \multicolumn{1}{c|}{78.0} & 40.2 & \multicolumn{1}{c|}{61.3} & 42.3 \\
STAR\cite{chen2024fine} & ACMM2024 & \multicolumn{1}{c|}{81.4} & 45.1 & \multicolumn{1}{c|}{63.3} & 44.3 \\
PURLS\cite{zhu2024part} & CVPR2024 & \multicolumn{1}{c|}{79.2} & 41.0 & \multicolumn{1}{c|}{72.0} & 52.0 \\
SA-DVAE\cite{li2024sa} & ECCV2024 & \multicolumn{1}{c|}{82.4} & 41.4 & \multicolumn{1}{c|}{68.8} & 46.1 \\
ScoPLe\cite{zhu2025semantic} & CVPR2025 &  \multicolumn{1}{c|}{84.1} & 53.0 & \multicolumn{1}{c|}{\textcolor{red}{74.5}} & 52.2 \\
Neuron\cite{chen2024neuron} & CVPR2025 &  \multicolumn{1}{c|}{\textcolor{blue}{86.9}} & \textcolor{blue}{62.7} & \multicolumn{1}{c|}{71.5} & 57.1 \\
TDSM\cite{} & ICCV2025 &  \multicolumn{1}{c|}{86.5} & 56.1 & \multicolumn{1}{c|}{74.2} & \textcolor{red}{65.1} \\
FS-VAE\cite{wu2025frequency} & ICCV2025 &  \multicolumn{1}{c|}{\textcolor{blue}{86.9}} & 57.2 & \multicolumn{1}{c|}{\textcolor{blue}{74.4}} & \textcolor{blue}{62.5} \\
\hline
\textbf{Ours} & \textbf{\textbackslash{}} & 
\multicolumn{1}{c|}{\textbf{\textcolor{red}{89.6}}} & 
\textbf{\textcolor{red}{64.4}} & 
\multicolumn{1}{c|}{74.2} & 
60.1 \\ \hline
\end{tabular}
}
\label{tab:zsl}
\end{table}

\begin{table}[htbp]
\scriptsize
\renewcommand\arraystretch{1.1}
\centering
\vspace{-10pt}
\caption{\textbf{Average ZSL and GZSL accuracy under the random split settings on the NTU-60 and PKU-MMD datasets.} SkeletonContext delivers stable and competitive results across datasets, reflecting robust generalization under varying unseen-class configurations.}
\vspace{-5pt}
\setlength\tabcolsep{4.0pt} 
{
\begin{tabular}{c|c|c|c|c|c|c|c|c}
\hline
Model & \multicolumn{4}{c|}{NTU-60 (55/5 split)} & \multicolumn{4}{c}{PKU-MMD (46/5 split)} \\
\cline{2-9}
& ZSL & Seen & Unseen & H & ZSL & Seen & Unseen & H \\
\hline
ReViSE\cite{hubert2017learning}       & 60.9 & 71.8 & 52.1 & 60.3 & 59.3 & 60.9 & 42.2 & 49.8 \\
JPoSE\cite{wray2019fine}        & 59.4 & 66.3 & 54.9 & 60.5 & 57.2 & 60.3 & 45.2 & 51.6 \\
CADA-VAE\cite{schonfeld2019generalized}     & 61.8 & \textcolor{blue}{77.4} & 58.1 & 66.4 & 60.7 & 63.2 & 35.9 & 45.8 \\
SynSE\cite{gupta2021syntactically}        & 64.2 & 75.8 & 60.7 & 67.5 & 63.1 & 49.6 & 40.5 & 49.5 \\
SMIE\cite{zhou2023zero}         & 65.1 & -    & -    & -    & 60.8 & -    & -    & -    \\
SA-DVAE\cite{li2024sa}      & 84.2 & \textcolor{red}{78.2} & 72.6 & 75.3 & 66.5 & 58.5 & \textcolor{blue}{51.4} & 54.7 \\
SCoPLe\cite{zhu2025semantic}       & 83.7 & 75.3 & \textcolor{red}{80.2} & \textcolor{red}{77.7} & \textcolor{blue}{71.4} & \textcolor{blue}{62.2} & 49.7 & 54.9 \\
Neuron\cite{chen2024neuron}       & \textcolor{blue}{84.5} & - & - & 71.2 & 70.6 & - & - & \textcolor{blue}{69.2} \\
\hline
\textbf{Ours}       & \textbf{\textcolor{red}{85.7}} & \textbf{\textcolor{red}{78.2}} & \textbf{\textcolor{blue}{73.4}} & \textbf{\textcolor{blue}{75.5}} & \textbf{\textcolor{red}{73.5}} & \textbf{\textcolor{red}{68.3}} & \textbf{\textcolor{red}{72.2}} & \textbf{\textcolor{red}{71.4}} \\
\hline
\end{tabular}}
\label{tab:zsl-gzsl-random}
\vspace{-0.3in}
\end{table}

\section{Experiments}
\subsection{Datasets}
We follow the evaluation protocol proposed in \cite{gupta2021syntactically} to assess the ZSSAR performance of our model across three benchmark datasets. Specifically, we report results using two evaluation metrics: (a) general performance, and (b) performance under three random class-split settings, also defined in \cite{gupta2021syntactically}. All splits are cross-subject, ensuring no overlap between subjects in the training and testing sets. In each evaluation split 'seen/unseen', which refers to the number of seen (training) classes and the number of unseen (test) classes, respectively.

\textbf{NTU-RGB+D 60}~\cite{shahroudy2016ntu} is a widely used benchmark for ZSSAR, containing 56,880 video samples across 60 daily actions. These actions are performed by 40 subjects and recorded from 80 distinct camera viewpoints. We utilize the skeleton data, which consists of 25 joints per person, with each sample including two performers. Following the zero-shot evaluation protocols in \cite{gupta2021syntactically}, we conduct experiments using 55/5 and 48/12 class splits.

\textbf{NTU-RGB+D 120}~\cite{liu2019ntu}, an extension of NTU-60, is currently one of the largest datasets available for ZSSAR. It comprises 114,480 video samples from 120 action classes, performed by 106 subjects and captured from 155 viewpoints. This dataset introduces significantly greater variability and complexity. In line with \cite{gupta2021syntactically}, we adopt the 110/10 and 96/24 class splits for evaluation.

\textbf{PKU-MMD}~\cite{liu2017pku} is a mid-scale dataset containing over 20,000 action instances across 51 action categories. It is particularly valuable for evaluating models on complex and interactive actions, including both individual and group behaviors. We compute the average accuracy over three random 46/5 splits, as used in \cite{li2024sa}.
\vspace{-5pt}

\subsection{Implementation Details}
We sample all skeleton sequences to a fixed length of 64 frames using the data preprocessing procedure described in \cite{chen2024neuron}. For the skeleton encoder, we adopt Shift-GCN \cite{cheng2020skeleton}, which is pretrained on the seen action categories. 
We use BERT-base-uncased as the pretrained language model for context reconstruction.
We employ ChatGPT-4 to generate 10 descriptions per action class for building the context prompts.
The model is optimized using Stochastic Gradient Descent (SGD) \cite{bottou2010large} with a weight decay of 0.0005. The batch size is set to 64, and the initial learning rate is 0.1, which is decayed by a factor of 0.1 at epochs 10 and 20. All experiments are conducted using the PyTorch framework on a system equipped with a GeForce RTX 3090 GPU.

\subsection{Comparison with State-of-the-Art}

We compare our method with several state-of-the-art zero-shot action recognition approaches under the same evaluation splits, and report the results of NTU-60~\cite{shahroudy2016ntu} and NTU-120~\cite{liu2019ntu} datasets in Table \ref{tab:zsl} and Table \ref{tab:gzsl}.
The results show that our proposed SkeletonContext achieves competitive performance across all benchmarks, demonstrating its strong capability to generalize from seen to unseen classes.
In particular for the more challenging GZSL task, our method achieves superior performance on the harmonic mean ($H=\frac{2 \times Seen \times Unseen}{Seen + Unseen}$), indicating a more balanced recognition between seen and unseen classes.
The consistent improvements across both setting demonstrate the strong generalization ability of our model in transferring knowledge from seen to unseen action categories.

The setting of class splits is crucial for accuracy calculation and Tables \ref{tab:zsl} and Table \ref{tab:gzsl} only show results of a few predefined splits, which can not infer the overall performance on a complete dataset. Thus, we follow previous approaches to randomly select several unseen classes as a new split, repeat it three times, and report the average performance.
We report the ZSL and GZSL results of NTU~\cite{shahroudy2016ntu,liu2019ntu} and PKU~\cite{liu2017pku} datasets in Table \ref{tab:zsl-gzsl-random}. It can be seen that our method consistently achieves state-of-the-art results in different settings, showing excellent stability and robustness.
\vspace{-0.1in}

\begin{table}[h]
\small
\renewcommand\arraystretch{1.0}
\centering
\caption{
\textbf{Performance comparison on similar classes under different difficulty levels.} SkeletonContext yields clear and consistent gains over state-of-the-art methods, especially for challenging (medium/hard) cases, validating its robustness in fine-grained action discrimination.}
\vspace{-4pt}
\begin{tabular}{@{}ccccccc@{}}
\toprule
\multirow{2}{*}{Methods} & \multicolumn{2}{c}{Easy Level} & \multicolumn{2}{c}{Medium Level} & \multicolumn{2}{c}{Hard Level} \\ \cmidrule(l){2-7} 
& ZSL& GZSL& ZSL& GZSL& ZSL& GZSL\\ \cmidrule(l){1-7}
Neuron\cite{chen2024neuron}&57.0&43.1&53.5&42.3&56.4&43.8\\
FS-VAE\cite{wu2025frequency}&61.5&55.1&60.5&54.5&54.3&50.7\\
Ours&63.3&56.8&62.1&57.8&56.7&55.8\\ 
$\Delta$&\textcolor{red}{\textbf{+1.8}}&\textcolor{red}{\textbf{+1.7}}&\textcolor{red}{\textbf{+1.6}}&\textcolor{red}{\textbf{+3.3}}&\textcolor{red}{\textbf{+2.4}}&\textcolor{red}{\textbf{+4.9}}\\ \bottomrule
\end{tabular}
\label{tab:difficulty}
\end{table}
\vspace{-0.4in}

\subsection{Performance on Similar Classes}
To further evaluate the ability of the model to distinguish actions with high inter-class similarity, we conducted experiments under different difficulty levels derived from the confusion results of FS-VAE~\cite{wu2025frequency} on the NTU120 \cite{liu2019ntu} (96/24 split). Specifically, for each unseen category, we construct a confusion matrix based on the’s recognition results of FS-VAE~\cite{wu2025frequency} and identify its top-3 most easily confused classes. Each unseen class is then taken as an anchor and grouped with its three most confusing counterparts to form an ambiguous cluster. These clusters are categorized into Hard, Medium, and Easy levels in ascending order of recognition difficulty.

As shown in Table \ref{tab:difficulty}, the performance of previous methods decreases dramatically as the difficulty level increases, indicating their limited ability to handle inter-class ambiguity.
In contrast, SkeletonContext maintains stable performance across all levels and achieves clear advantages under medium and hard settings.
This robustness stems from the Cross-Modal Context Prompt Module, which injects language, driven contextual semantics, such as interacted objects and environments, into the skeleton representations.
By leveraging these contextual cues, our model can capture fine-grained differences among visually similar motions.

\begin{table}[htb]
\small
\renewcommand\arraystretch{1.0}
\centering
\caption{
\textbf{Ablation on core modules of \textit{SkeletonContext}.} 
Incorporating the Differential Joint Encoder (DJE), Semantic Context Grounding (SCG), Progressive Partial Masking (PPM), and Key-Part Decoupling (KPD) modules progressively improves zero-shot transfer performance.
}
\label{tab:components}
\begin{tabular}{@{}cccccccc@{}}
\toprule
\multicolumn{4}{c}{Modules} & \multicolumn{2}{c}{\begin{tabular}[c]{@{}c@{}}NTU-60\\(55/5split)\end{tabular}} & \multicolumn{2}{c}{\begin{tabular}[c]{@{}c@{}}NTU-120\\(96/24split)\end{tabular}} \\
\midrule
DJE & SCG & PPM & KPD & ZSL & GZSL & ZSL & GZSL \\
\midrule
\ding{55} & \ding{55} & \ding{55} & \ding{55} & 79.4 & 68.8 & 52.6 & 49.4 \\
\checkmark & \ding{55} & \ding{55} & \ding{55} & 81.4 & 70.6 & 54.7 & 51.4 \\
\checkmark & \checkmark & \ding{55} & \ding{55} & 83.9 & 73.1 & 57.9 & 55.4 \\
\checkmark & \checkmark & \checkmark & \ding{55} & 87.4 & 75.2 & 59.2 & 55.9 \\
\checkmark & \checkmark & \checkmark & \checkmark & \textbf{89.6} & \textbf{77.1} & \textbf{60.1} & \textbf{56.1} \\
\bottomrule
\end{tabular}
\end{table}

\section{Ablation Study}

\noindent\textbf{Ablation on Core Modules.}
To verify the contribution of each proposed component, we conduct a series of ablation experiments on NTU-60 (55/5 split) and NTU-120 (96/24 split) datasets.
Starting from the baseline skeleton encoder, we progressively introduce the Differential Joint Encoder (DJE), Semantic Context Grounding (SCG), Progressive Partial Masking (PPM), and Key Part Decoupling (KPD) modules.
We observe that incorporating DJE noticeably enhances representation quality by explicitly modeling inter-joint differentials, which strengthens the encoding of spatial topology and motion semantics.
When SCG is further introduced, the model gains substantial improvement in cross-modal reasoning, demonstrating that injecting reconstructed contextual semantics effectively bridges the gap between skeleton and textual modalities.
Adding PPM refines the model’s ability to infer complete contextual cues from partial observations, enabling more stable context reconstruction during training.
Finally, integrating KPD further enhances discriminative focus on key motion parts, allowing the model to handle both context-dependent and context-independent actions in a unified manner.
Overall, these components contribute cumulatively and consistently, validating that each module plays a distinct yet complementary role in improving contextual reasoning and motion discrimination for zero-shot generalization.

\begin{table}[t]
\small
\renewcommand\arraystretch{1.0}
\centering
\begin{minipage}{.4\linewidth}
\caption{
\textbf{Effect of loss objectives in the \textit{SkeletonContext} framework.} 
Both $L_{ccr}$ and $L_{kpd}$ are guided by LLM-generated contextual semantics, and removing either degrades performance.
}
\label{tab:loss}
\setlength\tabcolsep{1.0pt}
\begin{tabular}{p{2cm} c c}
\toprule
\multirow{2}{*}{Methods} & \multicolumn{2}{c}{\begin{tabular}[c]{@{}c@{}}NTU-60\\ (55/5 split)\end{tabular}} \\
\cmidrule(r){2-3}
& ZSL & GZSL \\ 
\midrule
\textbf{Ours} & \textbf{89.6} & \textbf{77.1} \\
w/o $\mathcal{L}_{ccr}$ & 86.4 & 75.6 \\
w/o $\mathcal{L}_{kpd}$ & 88.2 & 76.1 \\
w/o $\mathcal{L}_{ccr}, \mathcal{L}_{kpd}$ & 83.2 & 75.1 \\
\bottomrule
\end{tabular}
\end{minipage}
\hspace{0.08\linewidth}
\begin{minipage}{.5\linewidth}
\caption{
\textbf{Effect of contextual slot design in the structured prompt.} 
Incorporating all three contextual slots, environment (EN), used object (UO), and target object (TO), achieves the best performance.
}
\label{tab:context}
\setlength\tabcolsep{3.0pt}
\begin{tabular}{@{}cccccc@{}}
\toprule
\multicolumn{3}{c}{Prompt Template} & \multicolumn{2}{c}{{\begin{tabular}[c]{@{}c@{}}NTU-60\\ (55/5 split)\end{tabular}}} \\ 
\cmidrule(lr){1-3} \cmidrule(l){4-5}
EN & UO & TO & ZSL & GZSL \\ 
\midrule
\checkmark & \ding{55} & \ding{55} & 84.4 & 74.8 \\ 
\ding{55} & \checkmark & \ding{55} & 85.6 & 75.1 \\ 
\ding{55} & \ding{55} & \checkmark & 85.1 & 75.4 \\ 
\checkmark & \checkmark & \ding{55} & 87.0 & 76.9 \\ 
\checkmark & \checkmark & \checkmark & \textbf{89.6} & \textbf{77.1} \\ 
\bottomrule
\end{tabular}
\end{minipage}
\vspace{-0.2in}
\end{table}

\noindent\textbf{Effect of Loss Objectives.}
We also analyze the impact of the two auxiliary objectives, namely the context reconstruction loss ($L_{ccr}$) and the key part decoupling loss ($L_{kpd}$) in Table \ref{tab:loss}.
Removing either term results in degraded performance, indicating their necessity for balanced learning. 
It is worth noting that the variant \emph{w/o $L_{ccr}$} effectively reflects the absence of LLM-based contextual supervision, since the context reconstruction process no longer receives semantic guidance from generated descriptions. 
Eliminating the LLM-provided contextual semantics (\emph{w/o LLM context generation}) led to an additional $\sim$2-3\% drop, demonstrating that language-informed context is essential for effective cross-modal alignment.

\noindent\textbf{Effect of Context Slot Design.}
We further investigate the contribution of different contextual slots in the structured prompt, as summarized in Table \ref{tab:context}.
The results reveal that the inclusion of object-related slots (UO and TO) provides a larger performance gain compared with the environment slot (EN) alone, indicating that interaction-related cues play a more decisive role in distinguishing fine-grained actions. This is intuitive, as many skeleton-based actions are primarily characterized by how the human body interacts with objects.
When all three slots are combined, the model achieves the best overall accuracy.
This confirms that integrating both environmental and object-centric contextual semantics yields the most comprehensive representation.

\subsection{Qualitative Analysis}

\begin{figure}[htb]
    \centering
    \includegraphics[width=\linewidth]{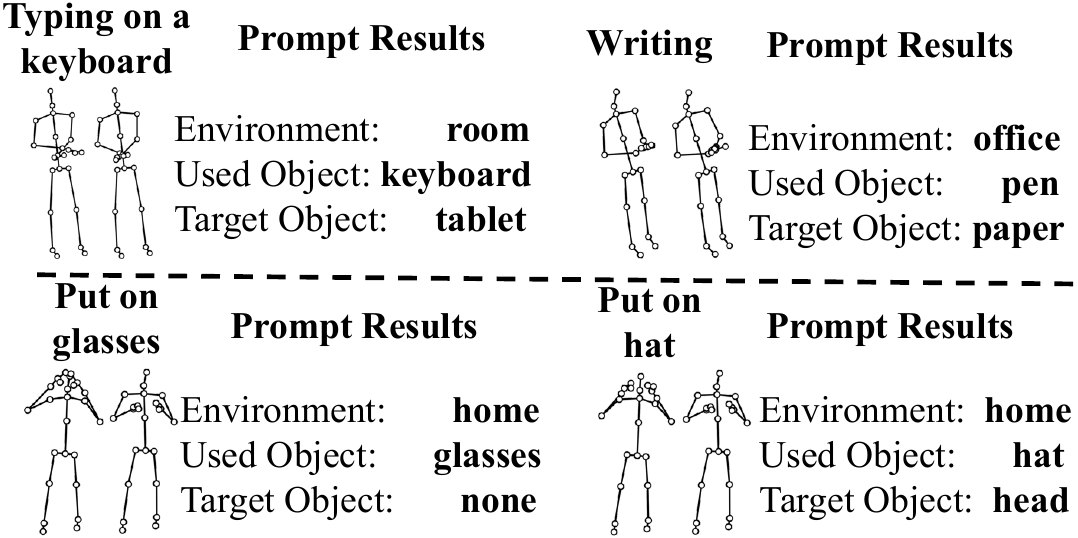}
    \caption{\textbf{Qualitative results of context reconstruction.} SkeletonContext can infer contextual semantics, enabling clear distinction between visually similar actions during inference.}
    \label{fig:vis_context}
\vspace{-0.1in}
\end{figure}

Fig.~\ref{fig:vis_context} illustrates that our method can accurately infer contextual semantics directly from skeleton data without any visual or textual input.
For instance, although \emph{typing on a keyboard} and \emph{writing} exhibit highly similar motion patterns, the model predicts distinct contextual elements (\emph{keyboard/tablet} vs. \emph{pen/paper}), enabling clear discrimination between these visually similar actions.
Similarly, for \emph{putting on glasses} and \emph{putting on a hat}, SkeletonContext correctly identifies the manipulated objects and corresponding target body parts, demonstrating its ability to associate motion dynamics with contextual semantics.

\begin{figure}[htb]
    \centering
    \includegraphics[width=\linewidth]{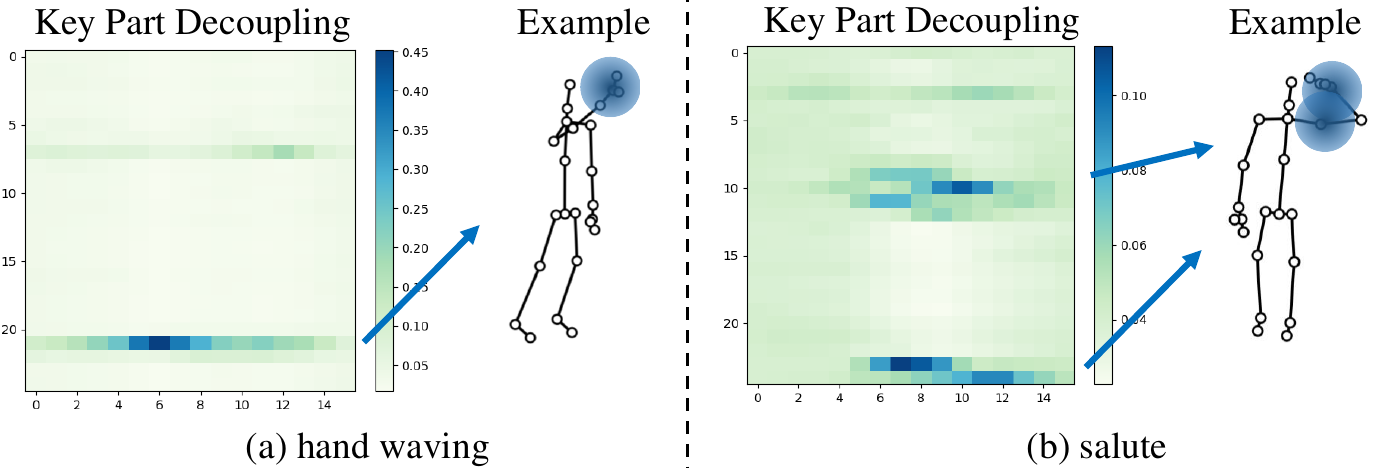}
    \caption{\textbf{Visualization of key-part decoupling.} The KPD module highlights motion-critical joints guided by language priors, revealing semantically relevant body parts for each action.}
    \label{fig:vis_part}
\vspace{-0.1in}
\end{figure}


Fig.~\ref{fig:vis_part} visualizes our method can highlight motion-critical joints guided by the language-informed priors. The heatmaps reveal that KPD consistently attends to semantically relevant body regions. For example, focusing on the hands for hand waving and the arm–head area for saluting.
These visualizations confirm that SkeletonContext not only reconstructs plausible contextual information but also grounds it to the most informative joints, achieving interpretable and context-aware zero-shot action recognition.
Additional visualizations, failure examples, and detailed analyses can be found in the supplementary material.

\section{Conclusion}
In this paper, we proposed SkeletonContext, a context-aware framework for zero-shot skeleton-based action recognition.
By introducing language-driven contextual reconstruction and key-part decoupling, our model effectively enriches skeleton representations with semantic context and highlights motion-critical joints, enabling robust generalization to unseen actions.
Qualitative results further show that SkeletonContext can accurately infer plausible contextual semantics, such as objects and environments, and localize the most informative body parts, demonstrating strong instance-level semantic grounding.
Extensive experiments across multiple benchmark datasets confirm that SkeletonContext achieves state-of-the-art performance under both ZSL and GZSL settings.
Moreover, evaluations on ambiguous class ZSL splits validate that our model can reliably distinguish visually similar actions by leveraging inferred contextual cues.
Future work will explore extending this framework to few-shot learning and real-world video understanding scenarios.

\medskip
\noindent\textbf{Acknowledgements.}  This work was supported by the Natural Science Foundation of Shanxi Province (Grant No. 2024JCJCQN-66) and the China Scholarship Council.

{
    \small
    \bibliographystyle{ieeenat_fullname}
    \bibliography{main}
}


\end{document}